\pdfoutput=1

\documentclass[11pt]{article}

\usepackage[final]{acl}
\usepackage{times}
\usepackage{latexsym}

\usepackage[T1]{fontenc}

\usepackage{listings}
\usepackage{alphabeta}
\usepackage{svg}

\usepackage[utf8]{inputenc}
\usepackage{microtype}
\usepackage{inconsolata}
\usepackage{graphicx}

\title{Opera Graeca Adnotata:\\ 

Building a 34M+ Token Multilayer Corpus for Ancient Greek}

\author{Giuseppe G. A. Celano \\
   Faculty of Mathematics and Computer Science\\
   Institute of Computer Science \\
Leipzig University \\
  \texttt{celano@informatik.uni-leipzig.de} }

\begin{document}
\maketitle
\begin{abstract}
In this article, the beta version 0.1.0 
of Opera Graeca Adnotata (OGA), the largest 
open-access multilayer corpus for Ancient Greek (AG) is presented. 
OGA consists of 1,687 literary works and 34M+ tokens coming 
from the \texttt{PerseusDL} and \texttt{OpenGreekAndLatin} 
GitHub repositories, which host AG texts ranging from 
about 800 BCE to about 250 CE.
The texts have been enriched with seven annotation 
layers: (i) tokenization layer; (ii) sentence segmentation layer; 
(iii) lemmatization layer; (iv) morphological layer; 
(v) dependency layer; (vi) dependency function layer; 
(vii) Canonical Text Services (CTS) citation layer.  
The creation of each layer is described by highlighting 
the main technical and annotation-related issues encountered. 
Tokenization, sentence segmentation, and CTS citation are performed
by rule-based algorithms, while morphosyntactic 
annotation is the output of the COMBO parser trained 
on the data of the Ancient Greek Dependency Treebank.
For the sake of scalability and reusability, the corpus 
is released in the standoff formats PAULA XML 
and its offspring LAULA XML.
\end{abstract}

\section{Introduction}

Multilayer corpora contain 
a variety of different annotations modeled as independent layers.
Contrary to corpora with inline annotations,
multilayer corpora have the unique advantage 
of \textit{scalability} 
and \textit{reusability}, in that a potentially infinite number of 
annotation layers can be added standoff,
which are connected to each other by references to base texts
in a graph structure.\footnote{
While \citet{zeldes2018multilayer} proposes a definition of
``multilayer'' with reference to independent 
annotation \textit{types},
I adopt a definition where independence simply refers
to formally separate standoff annotation 
layers, regardless of how content-wise independent layers are.}

An example of an open-access multilayer corpus for a modern
language is the National Corpus of Polish\footnote{\url{https://nkjp.pl/}} 
(\citealp{przepiorkowski2011national}),
 which is
encoded standoff according to the P5 TEI formalism:
Polish texts mostly coming from newspapers and magazines
are tokenized, sentence segmented, and annotated for
morphosyntax, named entities, and word sense disambiguation.\footnote{
Data can be downloaded 
at \url{https://clip.ipipan.waw.pl/NationalCorpusOfPolish}.
}

A number of historical language corpora have also
been annotated with different layers of linguistic information, 
such as 
Coptic Scriptorium\footnote{\url{http://copticscriptorium.org}} \cite{schroeder2016raiders} 
and RIDGES Herbology
\footnote{\url{https://www.laudatio-repository.org/browse/corpus/PySSCnMB7CArCQ9CNKFY/corpora}} 
\cite{odebrecht2017ridges},
which are both provided in standoff PAULA XML.
RIDGES Herbology, for example, contains
German herbal texts ranging from 1478 to 1870,
annotated with three different
transcription layers, i.e., a diplomatic one, which is
the closest to the original text,
and two normalization layers called ``clean'' 
and ``norm,'' respectively: the former aims 
to address the issue of a few character 
variations of the diplomatic transcription,
while the latter offers a higher-level uniformation
according to the principles of modern German
orthography. 

Besides morphosyntactic annotation,
it is noteworthy that the standoff nature 
of the RIDGES corpus allows addition of, for example, 
lexical layers, such as the one specifying 
a token's language
(e.g., whether it is in German or Latin) 
and the one containing the full name of
a person name, both of which would be difficult
to add to syntactic annotation inline. 

In the present article, 
I describe the beta version 
of Opera Graeca Adnotata (OGA),
a multilayer corpus for Ancient Greek (AG) 
\citep{oga}, 
focusing on its 
design, as well as on the technical and 
annotation-related issues encountered 
while dealing with a large-size dataset consisting of 
1,687 texts and 34,172,140 tokens, which
represents the largest 
open-access corpus so far published for AG.

The paper is organized as follows: 
in Section \ref{rw}, related work is presented,
while Section \ref{pxml} introduces the standoff 
formalism of PAULA XML and
its offspring LAULA XML. In 
Section \ref{pipe} and its subsections,
the texts (Section \ref{texts}) and their 
seven annotation layers are described:
(i) tokenization layer (Section \ref{toks}); 
(ii) sentence segmentation
layer (Section \ref{seg}); 
(iii) lemmatization layer (Section \ref{morph}); 
(iv) morphological layer (Section \ref{morph}); 
(v) dependency layer (Section \ref{morph}); 
(vi) dependency function layer (Section \ref{morph}); 
(vii) Canonical Text Services (CTS) citation layer 
(Section \ref{cts}).
Section \ref{concl}
concludes the article with a short summary 
and final remarks.

\section{Related Work}
\label{rw}

There exist a few noticeable corpora for AG. Thesaurus 
Linguae Graecae\footnote{\url{https://stephanus.tlg.uci.edu/}} is
 the largest non-open-access corpus for 
 AG (110M+ words); its texts
are accessible only via a query interface
with limited functionality (mostly word form and lemma search).
The open-access counterpart of TLG is represented
by Perseus Digital 
Library\footnote{\url{https://www.perseus.tufts.edu/hopper/}} 
and its offspring 
Scaife Viewer,\footnote{\url{https://scaife.perseus.org/}} 
whose collection of literary AG texts
coincides with that of OGA (see Section \ref{texts}):
both websites aim to offer a reading environment for
historical texts.

Among the ever-growing number of 
open-access resources, treebanks are 
particularly worth mentioning, 
in that they contain a variety of texts manually annotated for 
morphosyntax (see \citealp{celano2019dependency} 
for an overview), which many other resources, 
including OGA, (at least partly) rely on. 

The Diorisis corpus\footnote
{
\url{https://figshare.com/articles/dataset/The_Diorisis_Ancient_Greek_Corpus/6187256}} offers
lemmatization and morphological analyses 
of 820 texts and 10,191,371 tokens 
\cite{vatri2018diorisis}. More recently, the 
GLAUx project aims to
provide a larger morphosyntactically
and semantically annotated corpus, 
i.e., the GLAUx corpus \citep{keersmaekers2021glaux}:
at the moment, 
a demo version has been 
released,
\footnote{\url{https://github.com/perseids-publications/glaux-trees/tree/master/public/xml}} consisting 
of 13,374,342 morphosyntactically
annotated tokens.\footnote{More recently, an attempt
to annotate Ancient Greek more accutately is detailed in
\citet{Keersmaekers2023}, but no new data has yet 
been published.}

The above-mentioned
annotated corpora share the characteristic of being
encoded in a project-specific format, which
is meant to provide consumable, but hardly reusable, 
data.

\section{PAULA XML and LAULA XML}
\label{pxml}

PAULA XML (\textit{Potsdamer AUstauschformat 
Linguistischer Annotationen}) 
is an established open-access 
standoff format 
for linguistic annotation \citep{dipper2005xml}. 
It was inspired by LAF 
(Linguistic Annotation
Framework, \citealp{IDE_ROMARY_2004}), 
and therefore has many commonalities with
the non-open-access
GrAF format (Graph Annotation Format,
\citealp{ide2007graf}).

In PAULA XML, a base text is
directly or indirectly referenced by identifiers
contained in 
annotation layers, each of which
is stored in a separate file.\footnote{Standoff annotation
is typically performed by using separate files
for each annotation layer and indeed, this is
the PAULA XML model. However, it is to be noted that
standoff annotation is primarily defined by a 
referencing mechanism keeping markup data 
and text to markup separate, without this 
necessarily implying 
separation in different files.
However, for scalability purposes, different files
are typically used.}
All together,
the files form an acyclic graph.
A base text embeds the transcription
of an original text within a
shallow XML structure, so that it
can typically be referenced by 
at least one annotation layer,
i.e., a tokenization layer,
which identifies tokens
by referencing
character offsets.
Each thus identified token is
associated with an ID, i.e., a
unique identifier that can
in turn be referenced in 
other annotation layers. 

\begin{figure}[ht]
\includegraphics[width=\linewidth]{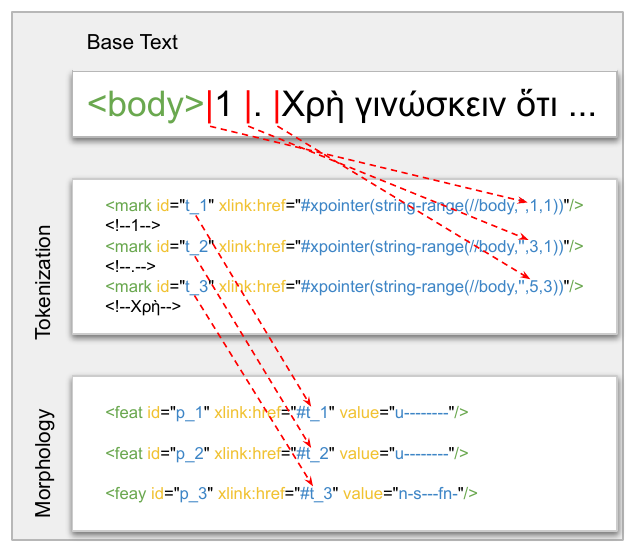}
\caption{Logic of standoff annotation layers in PAULA XML}
\label{fe2}
\end{figure}

Indeed, as shown in Figure \ref{fe2},
the XPointer expressions within 
the tokenization layer reference
the base text by specifying the
start offset of a token---numbering
starts with 1 and not 0---and
its length. The XPointer expressions
are associated with IDs, which
can then be used, for example, in the morphological
layer to associate each of them with
the corresponding morphological annotation
contained in the \texttt{value} attribute.

Such a model has the advantage of
offering an elegant solution to 
the issue of overlapping
markup. For example, a
layer for prosody annotation 
could refer to tokens 
different from
morphosyntactic tokens: for different
 tokenization layers to reference 
 the same base text would then guarantee
that their schemata
can be compared.

\begin{figure*}[ht]
\includegraphics[width=\linewidth]{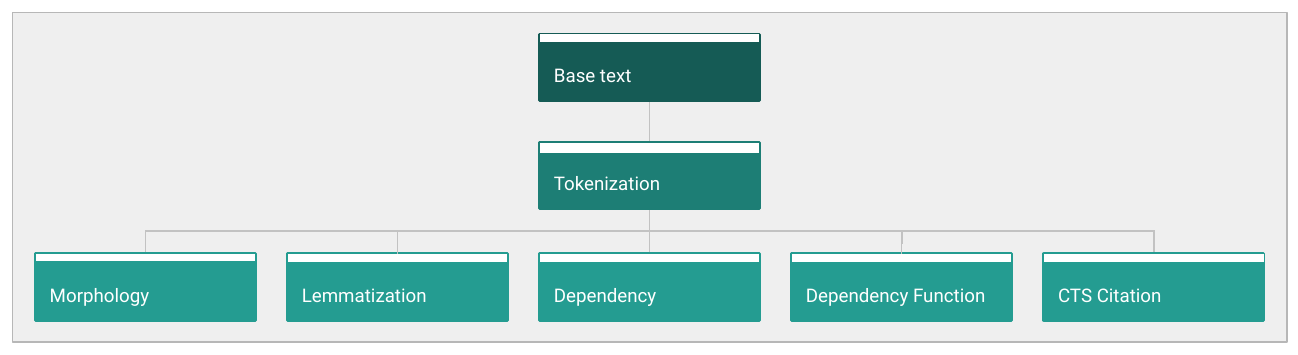}
\caption{Architecture of Standoff PAULA XML for OGA}
\label{feed}
\end{figure*}

Currently, OGA contains the following
annotation layers (see Figure \ref{feed}): (i) tokenization layer;
(ii) sentence segmentation layer;\footnote{
This layer is available in LAULA XML, but not in
PAULA XML (see Section \ref{seg}).
}
(iii) lemmatization layer; (iv) morphological layer;
(v) dependency layer;
(vi) dependency function layer;
(vii) CTS citation layer. 

Notably, PAULA XML
requires a base text to be inserted
in the \texttt{<body>} element of an XML file,
without any XML markup within it. This has an
impact for OGA, its main texts being extracted
from often complex EpiDoc TEI XML texts (see Section \ref{toks}).

PAULA XML is part of a set
of technologies developed for
annotation of multilayer corpora,
which includes ANNIS\footnote{\url{https://corpus-tools.org/annis/}
} \citep{krause2016annis3}, a query engine,
Salt, a meta model for manipulating
linguistics data, Pepper, a converter
between different annotation formats,
and Hexatomic, an annotation editor.

One disadvantage of PAULA XML 
is that the size of a corpus tends to
grow exponentially as new texts and/or
annotation layers are added. For example,
the directory containing the PAULA files 
of OGA 0.1.0 is
as large as about 15 GB (unzipped). 
For this reason, OGA files 
are processed using a
lighter and more efficient XML structure, called
LAULA XML (\textit{Leipziger AUstauschformat 
Linguistischer Annotationen}),
which retains the logic of PAULA XML,
but its repeating element and
attribute names are shortened to
one character (e.g., \texttt{<mark>} 
becomes \texttt{<m>} and
\texttt{<feature>} \texttt{<f>});
moreover, information that 
in PAULA XML can only be added inside
XML comments is 
conveniently encoded, in LAULA
XML, within XML attributes, whose content can usually be
retrieved much faster by XPath parsers.

LAULA XML also allows
original TEI XML files to be
directly referenced, without 
the need of modifying texts,
a particularly convenient
property for corpora such as
OGA, whose main texts only
consist of EpiDoc TEI XML files
(see Section \ref{texts}). In general,
PAULA XML is used 
as a serialization format for
the sake of ANNIS, the query engine
facilitating corpus query, while
LAULA XML has been 
created for a more efficient
XML parsing of the texts.

\section{The OGA pipeline}
\label{pipe}

Due to the high number of texts and
annotation layers to process, 
OGA is created automatically,
with minimal human inspection:
it is the output of 
a number of scripts, whose 
content is summarized 
in the following subsections.
Because of its large size, 
the corpus, a few
related resources, and its documentation 
is made available on Zenodo.\footnote{
\url{https://zenodo.org/records/8158675}
}

\subsection{The Texts}
\label{texts}

The main texts of OGA 0.1.0 come from
two GitHub repositories:
(i) Canonical Greek\footnote{
\url{https://github.com/PerseusDL/canonical-greekLit/?tab=readme-ov-file}
} and (ii) First1KGreek.\footnote{
\url{https://github.com/PerseusDL/canonical-greekLit/}} 
While the former contains
Classical Greek texts, which
mostly coincide with the ones available on
Perseus Digital Library 4.0,\footnote{
\url{https://www.perseus.tufts.edu/hopper/}
} the latter aims to complement
the former by adding one edition
of any Greek work composed until about 250 CE.

Since at least 2017, 
these texts have been edited actively for 
correction of OCR errors and, more in general,
for compliance with more recent standards.
In particular, there has been an ongoing effort to
transition older XML files into EpiDoc
P5 TEI XML files and provide each text
with CTS structure (see Section \ref{cts}). New releases
of both repositories are issued on a frequent
basis.

The texts in OGA 0.1.0 are a subset of those of
the releases \texttt{0.0.5217465242} (Canonical Greek)
and \texttt{1.1.5225573344} (First1KGreek), 
amounting to 1,687 texts out of
1,801. Indeed,
only CTS-compliant files are selected and
only the first file for a work is chosen, 
if, as occurs in a few cases, 
more than one file (i.e., edition) is
available.

Although correction of the original texts 
is outside the scope of OGA, some automatic
encoding normalization is performed because
characters with similar glyphs are often
confused and therefore inconsistently encoded 
with different Unicode codepoints.

More precisely, NFC normalization
is applied to address the issue of
characters such as ``epsilon with 
an acute accent,'' which can be encoded 
as the Unicode codepoint U+03AD or U+1F73, 
the former being in the ``Greek and Coptic'' 
chart, while the latter in the ``Greek Extended'' one:
only the codepoint U+03AD is, however,
the NFC-normalized codepoint, and therefore
NFC normalization will make encoding of this
character uniform. 

NFC-normalization cannot, however, solve all
character inconsistency--related issues.
For this reason, an \textit{ad-hoc} script tries to normalize
the encoding of the rather frequently
occurring apostrophe character, for which---as 
in the encoding of other languages---different 
Unicode codepoints tend to be used.

More precisely, the
codepoints COMBINING COMMA ABOVE (U+0313),
GREEK KORONIS (U+1FBD), 
APOSTROPHE (U+0027), and RIGHT
SINGLE QUOTATION MARK (U+2019)
are substituted with
MODIFIER LETTER APOSTROPHE (U+02BC).
Since the APOSTROPHE (U+0027) and RIGHT
SINGLE QUOTATION MARK (U+2019) at the end of
a word cannot 
be unambiguously identified as representing
an apostrophe, a word list 
for the most common elided Greek words
is used to this end, which is also made available in
the corpus release.

\subsection{Tokenization}
\label{toks}

Since the original texts 
are encoded as EpiDoc TEI XML texts,
preprocessing is needed to separate
the text of a work, which
annotation layers refer to, from its paratext,
which is not annotated. 
This is a non-trivial task with
heavily marked-up historical texts,
because the distinction between text and paratext
is signalled in TEI documents via use
of many different XML elements that
serve different functions.

For example, the element \texttt{<note>},
as the name itself suggests,
contains a note, and therefore its content can be
safely identified as paratext. Similarly,
the contents of elements such as 
\texttt{<app>} and \texttt{<bibl>} can be discarded,
in that they unambiguously identify
paratext related to critical apparatus and
bibliography, respectively. 

The content of other TEI elements is, however, 
part of a text:
for example, \texttt{<foreign>} contains 
a foreign language term, while \texttt{<add>}
signals a text addition by
an editor, which should arguably be considered
as part of the main text.

In a few cases, the semantics, and consequently
the structure, of a TEI element is complex.
For example, \texttt{<choice>} presents
a number of alternative readings for 
a specific passage: it can contain
\texttt{<sic>} to highlight a certain
word form whose correction is
given in its sibling element \texttt{<corr>};
or it can contain the sibling elements  \texttt{<abbr>} and
\texttt{<expan>} for 
an abbreviation and its expansion, respectively.

Currently, OGA contains 
a morphosyntactic tokenization of the texts 
performed rule-based, in that there is 
almost a one-to-one correspondence between 
graphic words and morphosyntactic
tokens. The most notable
exception to this occurs
in case of crasis, i.e., the phonological
phenomenon whereby two words can be univerbated:
for example, the word κἐκεῖνος consists of
the words καὶ (``and'') and ἐκεῖνος (``that''). 
Since they belong to different POS, they need to
be separated in a morphosyntactic
tokenization schema. Luckily, most cases of crasis
can be unambiguously identified 
because of the punctuation
mark ``coronis'' (i.e., the 
Unicode codepoint COMBINING COMMA ABOVE, U+0313)
placed on a non-initial vowel. 

On the basis of this formal criterion,
the texts have been searched for 
words with a coronis and a list of them,
which is also made available in the OGA 
 release, 
has been compiled and used for tokenization.
Cases where a coronis is
more difficult to identify,
as when a smooth breathing is on
a word-initial vowel, 
have been left untreated.

Notably,  
XPointer expressions in a 
PAULA XML tokenization layer
reference a base text that only consists
of a long string
contained in a \texttt{<body>} element
(see Figure \ref{fe2}), in that no XML markup is
allowed in it;
on the other hand, since base texts 
in LAULA XML
coincide with the original EpiDoc TEI XML texts,
tokenization layers can reference
them conveniently through XPath expressions 
that follow their CTS structures (see Section \ref{cts}). 

\begin{table*}[ht]
\centering
\begin{tabular}{lllllll}
\hline
\textbf{Treebank} & \textbf{UPOS} & 
\textbf{XPOS} & \textbf{FEATS} & \textbf{LEMMA} &
\textbf{UAS} & \textbf{LAS}\\
\hline
UD Ancient Greek Perseus & 91.78 & 72.33 & 90.24 & 84.65 & 80.27
& 74.20\\
AGDT 2.1 & 94.04 & 85.41&89.48 & 88.95& 71.47&63.55  \\
\hline
\end{tabular}
\caption{\label{udagdt} Comparison of
F1 scores for the COMBO parser trained on 
the AGDT 2.1 texts and on their partial 
conversion into the UD annotation scheme 
(the UD scores are from 2018 Shared Task).}
\end{table*}

\subsection{Sentence Segmentation}
\label{seg}

Similarly to tokenization,
sentence segmentation is achieved in OGA through 
a rule-based algorithm. In fact,
sentence boundaries
are signaled unambiguously in
modern editions of AG texts
by the period, semicolon,
and middle dot punctuation marks.

The algorithm also addresses the issue
of use of parentheses---mostly
of editorial meaning---in 
conjunction with other sentence-final 
punctuation
marks, in that their order is not
standardized, but usually 
follows a modern language's 
style rules.

The sentence segmentation layer is
made available only in LAULA XML; in fact,
as explained in Section \ref{texts}, 
PAULA XML is mainly used as a serialization format 
for ANNIS, which can display and query 
annotations without the help of
sentence boundaries.

\subsection{Morphosyntactic Annotation}
\label{morph}

The morphosyntactic annotation in 
OGA 0.1.0 is performed automatically 
using the COMBO
parser,\footnote{
\url{https://github.com/360er0/COMBO}
} trained on the data of the
Ancient Greek Dependency Treebank 2.1
(AGDT).

COMBO is a jointly trained
POS tagger, lemmatizer, and dependency parser,\footnote{For
brevity's sake, I just call it ``parser.''} 
which participated in the 2018 CoNLL Universal 
Dependency Shared Task and ranked among the best
parsers (3-5 for LAS, 4 for MLAS, and
3-4 for BLEX).\footnote{
\url{https://universaldependencies.org/conll18/results.html}
} COMBO also scored among the best systems when
specifically evaluated on the AG
data, which derive from
AGDT 2.1: more precisely, the UD conversion contains
a subset of the data (approximately $2/5$ of the 
sentences/tokens contained in AGDT 2.1).

Differently from other parsers, COMBO is
open access, easy to use and, most importantly,
can also be trained on treebanks 
whose annotation schema is similar to---but does 
not coincide with---that of Universal Dependency (UD).

When trained on the original AGDT data,
COMBO returns F1 scores for UPOS,
FEATS, and LEMMA
comparable to the 2018 Shared Task ones; however,
the scores for XPOS, UAS, and LAS differ
by about 10.8\% on average (see Table 
\ref{udagdt}).

The AGDT annotation scheme derives from that
of the Prague Dependency 
Treebank\footnote{\url{https://ufal.mff.cuni.cz/prague-dependency-treebank}} 
\cite{11234/1-2621} and
consists of four annotation layers:
(i) morphological layer;
(ii) lemma layer;
(iii) dependency layer;
(iv) dependency function layer.

The morphological annotation
is represented as a 9-character
string, where the first character
is the part of speech
of a token and the remaining characters its morphological
features. The lemma annotation
refers to the
dictionary entry for a token: remarkably,
the AGDT annotation schema follows
the convention of representing
lemmas as 
\textit{single} word forms, even
if, in traditional grammar, lemmas
consist of more word forms, which
describe a token more accurately: for example, 
a lemma for a noun in an AG dictionary consists 
not only of its nominative form, as in  AGDT,
but also of its genitive, which conveys
relevant information about its declension.

The syntactic annotation follows a dependency
formalism, which consists in the identification of 
directed relations between
a head token and its dependent token---with a head 
token being able to have more
than one dependent, but not viceversa. 
When considered together, all relations 
form a directed acyclic graph,
more commonly described as an (upside-down) 
syntactic tree. Each
syntactic dependency is also typed with a syntactic label
expressing the function a dependent
has with reference to its head (e.g., 
subject or attribute).

Dependency grammar formalism is quite popular in
computational linguistics because it represents
a balanced trade-off between syntactic analysis 
precision and annotation feasibility. 
Describing the many details of the syntactic annotation is
outside the scope of the article: 
 the reader 
is referred for them to \citet{celano2019dependency}
and the annotation guidelines 
(\citealp{Craneguid}; \citealp{Celanoguid}).

\subsection{CTS Citation}
\label{cts}

CTS citation 
refers, in general, to a protocol
to retrieve passages from a literary work
by means of URNs \citep{blackwell2020cite},
which include identifiers for
an author, work, edition, and passage.

In reference to the annotation layer, however,
the phrase ``CTS citation'' 
is used with a narrower scope,
just to indicate a tag for the passage
assigned to each
token.\footnote{Actually, an identifier for an 
author, work, edition is contained
in the file name of each original
text file, and is retained also in the
OGA file names.}
For example, since Herodotus' Histories 
are
divided into
books, chapters, and sections, a CTS citation
tag provides each token of this 
work with
the number of the book, chapter, and section
it belongs to.

Indeed, OGA base texts are based only on those 
EpiDoc TEI XML texts containing
a \texttt{<refsDecl n="CTS">} element (within
the \texttt{<encodingDesc>} element),
in which an XPath expression is
provided for identification of
work divisions. For example,
in the file \texttt{tlg1600.tlg001.perseus-grc2.xml}
(corresponding to \textit{Flavii Philostrati Opera}, Vol. 2),
the following XPath expression is given:
\begin{verbatim*}
/tei:TEI/tei:text/tei:body
/tei:div/tei:div[@n="$1"]
/tei:div[@n="$2"])
\end{verbatim*}
. The variables \texttt{\$1} and
\texttt{\$2} stand, respectively, 
for the numbers 
of the first and second kind of
division (\texttt{div}),
which, in this case, correspond to 
``book'' and ``chapter''---the 
division names are also
specified within a \texttt{<refsDecl>} element.

The importance of the CTS citation layer
for philological, historical, and linguistic 
studies cannot be overstated, considering how 
heavily they rely on the network of references made possible
through such passage numbering system.

\section{Conclusion and Outlook}
\label{concl}

In this paper, the architecture
of the beta version 0.1.0 of 
the OGA corpus has been presented. The base 
texts and their seven annotation layers 
have been described: (i) tokenization layer; (ii)
sentence segmentation layer; (iii) lemmatization
layer; (iv) morphological layer; (v) dependency 
layer; (vi) dependency function layer; (vii) CTS
citation layer. The layers are 
serialized as standoff PAULA XML and 
standoff LAULA XML for maximum 
scalability and reusability.

I examined the issues and challenges related 
to character encoding normalization and
to the rule-based tokenization of base texts, which
are often extracted from 
complex EpiDoc TEI XML files. The main
elements of the automatic morphosyntactic annotation
and CTS citation layer
have also been outlined. 

A stable version
of OGA is planned to be released 
with an evaluation of 
the performance of the tokenizer
and sentence segmenter and 
 with a refinement of the morphosyntactic
annotation. Other annotation layers,
such an IPA transcription one, are
also planned to be added.

\section{Acknowledgements}

This work has been supported by the German 
Research Foundation (DFG project number 408121292).


\end{document}